\documentclass[iicol,pdflatex]{sn-jnl} 

\usepackage{graphicx}%
\usepackage{multirow}%
\usepackage{amsmath,amssymb,amsfonts}%
\usepackage{amsthm}%
\usepackage{mathrsfs}%
\usepackage[title]{appendix}%
\usepackage{xcolor}%
\usepackage{textcomp}%
\usepackage{manyfoot}%
\usepackage{booktabs}%
\usepackage{algorithm}%
\usepackage{algorithmicx}%
\usepackage{algpseudocode}%
\usepackage{listings}%
\usepackage{arydshln}

\usepackage{natbib}

\usepackage{pifont}
\newcommand{\xmark}{\ding{55}} 
\newcommand{\cmark}{\ding{51}} 

\def\method{WildIng}

\theoremstyle{thmstyleone}%
\theoremstyle{thmstyletwo}%

\theoremstyle{thmstylethree}%

\begin{document}

\title{WildIng: A Wildlife Image Invariant Representation Model for Geographical Domain Shift}

\author*[1,2]{\fnm{Julian D.} \sur{Santamaria}}\email{julian.santamaria@udea.edu.co}  
\author[1]{\fnm{Claudia} \sur{Isaza}}\email{victoria.isaza@udea.edu.co}  
\author[2]{\fnm{Jhony H.} \sur{Giraldo}}\email{jhony.giraldo@telecom-paris.fr}  

\affil[1]{\orgdiv{SISTEMIC, Faculty of Engineering}, \orgname{Universidad de Antioquia-UdeA}, \orgaddress{\city{Medellín}, \country{Colombia}}}  
\affil[2]{\orgdiv{LTCI, Télécom Paris}, \orgname{Institut Polytechnique de Paris}, \orgaddress{\city{Palaiseau}, \country{France}}}








\abstract{
Wildlife monitoring is crucial for studying biodiversity loss and climate change. Camera trap images provide a non-intrusive method for analyzing animal populations and identifying ecological patterns over time. 
However, manual analysis is time-consuming and resource-intensive.
Deep learning, particularly foundation models, has been applied to automate wildlife identification, achieving strong performance when tested on data from the same geographical locations as their training sets. 
Yet, despite their promise, these models struggle to generalize to new geographical areas, leading to significant performance drops. For example, training an advanced vision-language model, such as CLIP with an adapter, on an African dataset achieves an accuracy of 84.77\%. However, this performance drops significantly to 16.17\% when the model is tested on an American dataset. 
This limitation partly arises because existing models rely predominantly on image-based representations, making them sensitive to geographical data distribution shifts, such as variation in background, lighting, and environmental conditions.
To address this, we introduce \method, a \textbf{Wild}life image \textbf{In}variant representation model for \textbf{g}eographical domain shift. \method~integrates text descriptions with image features, creating a more robust representation to geographical domain shifts. 
By leveraging textual descriptions, our approach captures consistent semantic information, such as detailed descriptions of the appearance of the species, improving generalization across different geographical locations.
Experiments show that \method~enhances the accuracy of foundation models such as BioCLIP by 30\% under geographical domain shift conditions.
We evaluate \method~on two datasets collected from different regions, namely America and Africa.
The code and models are publicly available at \url{https://github.com/Julian075/CATALOG/tree/WildIng}.
}




\maketitle

\section{Introduction}\label{intro}

Camera trap images are one of the most valuable data sources for wildlife monitoring, playing a crucial role in biodiversity conservation and climate change research \citep{reynolds2024potential,gadot2024crop,giraldo2019camera}. 
These images provide a non-intrusive and scalable way to study animal populations, track endangered species, and understand ecological patterns over time \citep{pollock2025harnessing,li2022cameratrapr,santamariaaudio}. 
By capturing images in remote locations, camera traps allow researchers to collect extensive datasets without direct human intervention, making them an essential tool for ecological studies. Considering the vast volume of collected images, it is imperative to implement automatic techniques for the identification of animal species present within the images.

With the rise of large-scale deep learning models, researchers have started exploring the use of Foundation Models (FMs) in wildlife monitoring \citep{yang2025enhancing,gabeff2024wildclip,fabian2023knowledge}.
FMs are trained on vast and diverse datasets, sometimes containing billions of data samples, allowing them to learn rich and transferable representations \citep{tang2025towards,yang2025learning,wu2024transferring}.
These models have demonstrated remarkable performance across various computer vision tasks, including image classification, object detection, and semantic segmentation, proving their flexibility and adaptability to different applications \citep{luo2024towards,riz2024novel,zang2024contextual}.

Recently, researchers have begun adapting FMs for camera trap image recognition. 
Instead of training models from scratch, current approaches aim to fine-tune \citep{yang2025enhancing} or adjust pre-trained FMs to incorporate domain-specific knowledge \citep{fabian2023knowledge}.
Some methods introduce adapters, which allow models to specialize in camera trap images without losing their general knowledge \citep{pantazis2022svl}.
Other models apply \textit{learning without forgetting} strategies, ensuring that models retain their broad capabilities while improving performance on wildlife images \citep{gabeff2024wildclip}.
Additionally, some approaches leverage external knowledge sources, such as internet databases, to refine the model’s understanding of specific animal species and their attributes \citep{fabian2023knowledge}. 
These strategies aim to bridge the gap between the general-purpose knowledge of FMs and the specialized needs of camera trap image recognition.

Despite their impressive performance with in-domain geographical data (data coming from the same geographical locations), these FM-based approaches often struggle when tested on out-of-domain geographical data (training and test data come from different geographical locations) \citep{hogeweg2024cood,norman2023can,tuia2022perspectives}.
This limitation is particularly problematic for camera trap applications, where the geographical locations differ substantially from those seen during the training phase \citep{schneider2020three, beery2018recognition, villa2017towards}.

\begin{figure}
    \centering
    \includegraphics[width=\columnwidth]{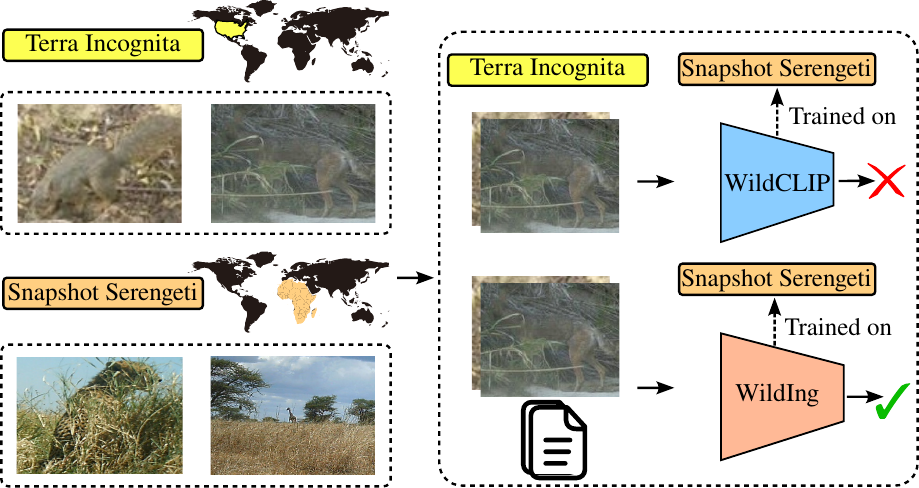}
    \caption{Comparison of \method~and WildCLIP \citep{gabeff2024wildclip} under geographical domain shift. Both models are trained on the Snapshot Serengeti dataset from Africa \citep{dryad_5pt92} and evaluated on the Terra Incognita dataset from the United States \citep{beery2018recognition}. \method~ demonstrates superior performance.}
    \label{fig:teaser}
\end{figure}

We observe that incorporating text into the input representation for camera trap images helps extract stronger features, alleviating the geographical domain shift issue.
In contrast, current models depend only on visual features, which are highly sensitive to changes in data distribution \citep{fang2025out,yu2023distribution}. Furthermore, many of these models are built on CLIP \citep{radford2021learning}, which has shown a tendency to lose its generalization ability \citep{wang2025attention,li2025clip} and become more susceptible to spurious correlations \citep{wang2024sober,kempf2025and} when the model is fine-tuned. As a result, CLIP-based models (e.g., WildCLIP \citep{gabeff2024wildclip} and BioCLIP \citep{stevens2024bioclip}) that rely solely on visual features often struggle to recognize images correctly in new geographical locations.  
Figure \ref{fig:teaser} provides an example of these observations, showing how such geographical variations cause the model’s learned features to fail in generalizing effectively, leading to misclassifications \citep{liang2023accuracy,wald2021calibration}.

In this paper, we introduce the \textbf{Wild}life image \textbf{In}variant representation model for \textbf{g}eographical domain shift (\method). 
Our approach introduces a simple yet effective new representation for wildlife monitoring data to address geographical domain shifts. 
This representation consists of using visual features in addition to text descriptions about the input images. This approach allows the model to capture geographical domain-invariant features by leveraging text descriptions, which remain consistent across different geographical regions. 
\method~consists of three main components: a text encoder, which includes a Large Language Model (LLM); an image encoder; and an image-text encoder, which incorporates a Vision-Language Model (VLM) and a Multi-Layer Perceptron (MLP). The MLP is used to address the domain shift between encoders, caused by the different feature spaces introduced by the VLM and LLM components \citep{duan2022multi}. An overview of the architecture is provided in Section~\ref{sec: overview} and illustrated in Figure~\ref{fig:pipeline}.

To evaluate \method, we train our model on one dataset and test it on another dataset from a different geographical region.
This setup allows us to analyze how well the model adapts to new environments where differences in background, lighting, and species composition create geographical domain shifts.
Our results show that \method~either outperforms or achieves competitive performance compared to general-purpose and domain-specific FMs in camera trap image recognition, particularly when the training and testing distributions differ due to geographical variations.

In this work, we build upon and improve our preliminary study \citep{Santamaria_2025_WACV}.
To achieve this, we introduce modifications to the model architecture and perform additional experiments.
More specifically, we replace the backbone of our model by changing the combination of CLIP (Contrastive Language-Image Pre-Training) \citep{radford2021learning} and BERT (Bidirectional Encoder Representations from Transformers) \citep{devlin2019bert} with Long-CLIP \citep{zhang2024long}.
Additionally, we modify the class representation, using only the information provided by LLMs.
We also evaluate \method's robustness to multiple random initializations to analyze the effect of the introduced changes on its performance.
Furthermore, we add new baselines for comparison.
Finally, we perform additional ablation studies and sensitivity analyses to provide a deeper understanding of the contribution of each component in our approach.

In summary, our main contributions are: 
\begin{itemize}[1.]
    \item  We introduce a novel \method~model to represent wildlife monitoring data, enhancing the extraction of geographical domain-invariant features. 
    \item When tested on datasets that differ from its training data, \method~outperforms previous FMs in recognizing animal species from camera trap images.
    \item We conduct a series of ablation studies to validate the effectiveness of each component in our model.
\end{itemize}

\section{Related Work}
\label{sec:related_works}
\subsection{Foundation Models}
In recent years, FMs have emerged as a new approach that achieves remarkable performance across a wide range of tasks without requiring task-specific training. These models leverage large-scale pre-training to learn high-level representations, leading to significant advancements in the field of machine learning \citep{awais2025foundation,huang2024language,jiang2023mistral,touvron2023llama}.
A key advancement in this field was CLIP \citep{radford2021learning}, which introduced a new learning approach by aligning visual features with text descriptions. 
CLIP significantly improved generalization across different tasks.
Later models, such as Long-CLIP \citep{zhang2024long}, extend the sequence length for better contextual understanding. 
Furthermore, CLIP-Adapter \citep{gao2024clip}, which refines CLIP’s learned representations using lightweight adaptation layers, continues to improve the alignment between visual features and text descriptions.
More recently, LLMs and VLMs have demonstrated strong capabilities in processing and generating both textual and visual content \citep{zhang2024vision,abdin2024phi,pmlr-v202-li23q}. Examples include GPT-4 \citep{achiam2023gpt} and LLaVA \citep{liu2024visual}, which leverage large-scale datasets to improve language and vision understanding across various applications.

\subsection{Foundation Models for Biology} 
FMs have been adapted to address domain-specific challenges, particularly in biological research, where data is often complex and specialized. Most of the adaptations of FM in biology are related to processing text, extracting biological information \citep{jung2024llm,lam2024large}, and modeling biological structures \citep{garau2025multi,jumper2021highly}. 
Beyond language processing and structural modeling, FMs have also been applied to vision-based biological tasks. One example is BioCLIP \citep{stevens2024bioclip}, which extends the principles of CLIP \citep{radford2021learning} to biological data, enabling the classification of diverse categories such as plants, animals, and fungi. 
Unlike general-purpose vision-language models, BioCLIP integrates structured biological knowledge and leverages taxonomic information, improving performance in fine-grained classification tasks \citep{stevens2024bioclip}.

\subsection{Foundation Models for Camera Trap Images}

The adaptation of FMs has extended beyond general and biological applications to camera trap image recognition, where they play a crucial role in wildlife monitoring and conservation.
One such model is WildCLIP, which leverages CLIP’s ability to align visual features with text descriptions to accurately classify animal species in camera trap images \citep{gabeff2024wildclip}. 
Similarly, WildMatch introduces a zero-shot classification framework by generating detailed visual descriptions of camera trap images and matching them to an external knowledge base for species identification \citep{fabian2023knowledge}. 
Another approach, Eco-VLM, enhances models that align visual features with text descriptions for ecological applications by fine-tuning on wildlife-specific datasets and applying text augmentation techniques \citep{yang2025enhancing}.
In contrast to models that align visual features with text descriptions, a more traditional deep learning approach was proposed by \cite{gadot2024crop}, who explored large-scale training for EfficientNetV2-M \citep{tan2021efficientnetv2}, a CNN-based architecture.

While previous methods have significantly improved camera trap image recognition in geographically in-domain evaluation, they still struggle when applied to different geographical regions and unseen species \citep{zhu2024vision,simoes2023deepwild,gadot2024crop}.
To address this limitation, our proposal introduces a more robust representation for wildlife monitoring data. It leverages detailed descriptions from a VLM to incorporate semantic invariant features, which are then used together with image features. Furthermore, the inclusion of more detailed class descriptions generated by the LLM improves the alignment between the input representation and the corresponding class.
This approach improves the input representation, enhancing robustness to geographical domain shifts.

\section{\method}
\label{sec:Method}

\subsection{Problem Definition}

The objective of this paper is to train a model in an annotated dataset of camera trap images from a specific geographical location, denoted as $\mathcal{D}$, which consists of $N_d$ image-label pairs, $\mathcal{D}=\{(\mathbf{x}_{i}^{D},\mathbf{y}_{i}^{D})\}_{i=1}^{N_d}$, with a set of classes $\mathcal{C}^{D}$.
Therefore, we evaluate the model's performance on a different camera trap image dataset from another geographical location, denoted as $\mathcal{S}$, which represents a distinct geographical domain, containing $N_s$ image-label pairs, $\mathcal{S} = \{ (\mathbf{x}_{i}^{S},\mathbf{y}_{i}^{S}) \}_{i=1}^{N_s}$, with a set of classes $\mathcal{C}^{S}$.
The set of classes of both datasets may or may not overlap, meaning that $\mathcal{C}^{D} \cap \mathcal{C}^{S}$ may or may not be empty.
Both datasets are derived from the natural world, but their image distributions differ due to being collected from different geographical regions, as illustrated in Figure \ref{fig:teaser}.
Our goal is to train a deep learning model using only the training dataset $\mathcal{D}$ and deploy it on the testing dataset $\mathcal{S}$.
This setting is highly practical in camera trap image research because the data used for testing usually comes from a different geographical domain than the training data.

\begin{figure}
    \centering
    \includegraphics[width=\columnwidth]{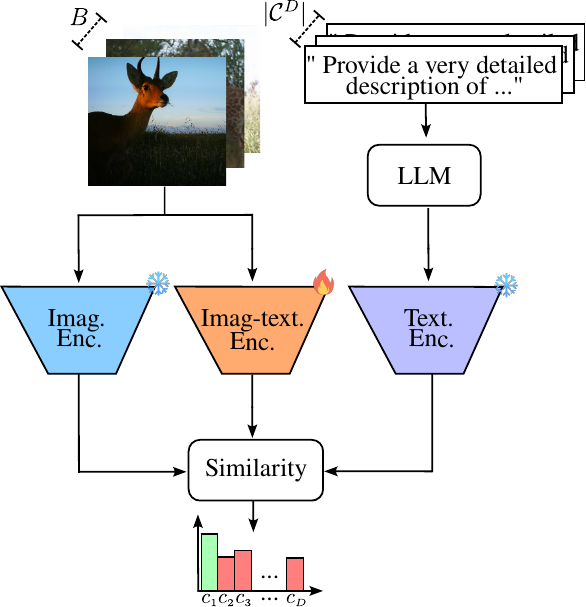}
    \caption{Overview of \method. The model integrates image, text, and image-text encoders along with an LLM. By leveraging text descriptions and image features, it extracts invariant features, improving robustness against geographical domain shifts.}
    \label{fig:pipeline}
\end{figure}

\subsection{Overview of the Approach}
\label{sec: overview}
The architecture of our proposed model, \method, is illustrated in Figure \ref{fig:pipeline}.
It consists of three main components: i) text encoder, ii) image encoder, and iii) image-text encoder.
In the text component, \method~uses an LLM to extract class-specific knowledge for each category in our dictionary of classes, $\mathcal{C}^{D}$. 
Then, the LLM-generated descriptions are processed by the text encoder. The resulting text embeddings are used to compute class-specific centroids for each class in $\mathcal{C}^{D}$ (Section~\ref{sec:Text Embeddings}).
This process produces a single embedding of dimension $F$.
For the image component, the model uses the image encoder to extract embeddings from a mini-batch of $B$ images (Section~\ref{sec:Image Embeddings}).
In the image-text encoder, \method~employs a VLM coupled with a text encoder and an MLP to compute image-text embeddings from the mini-batch of images (Section~\ref{sec:Image-text Embeddings}).
Text, image, and image-text embeddings are matched using a similarity mechanism (Section~\ref{sec:embeddings_aligmnet}). 
Finally, we utilize the output of the similarity mechanism to compute a contrastive loss, which is used to train our model (Section~\ref{sec:contrative}).
Most modules in \method~are frozen (\raisebox{-.3ex}{\includegraphics[height=0.3cm]{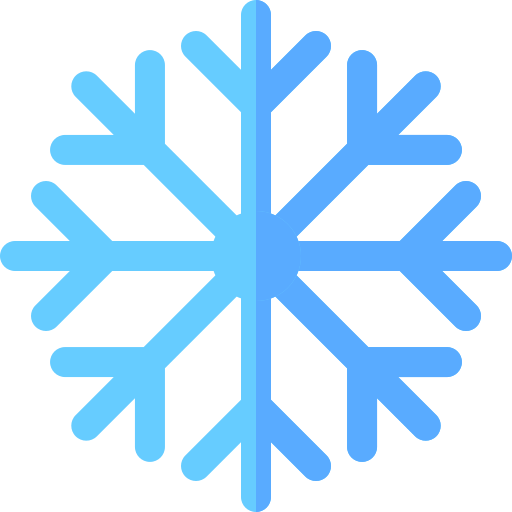}}) apart from the image-text encoder (\raisebox{-.3ex}{\includegraphics[height=0.3cm]{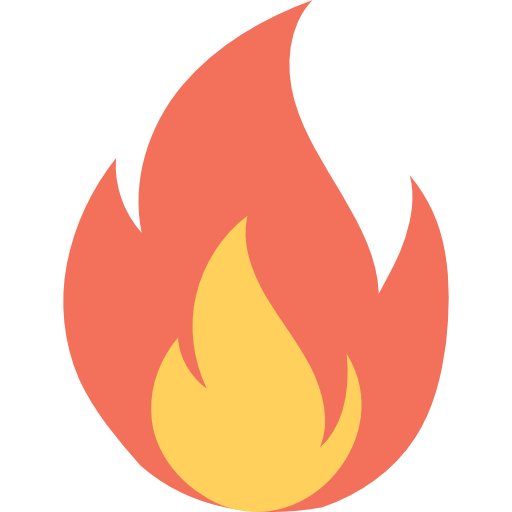}}).


\subsection{Text Encoder}
\label{sec:Text Embeddings}

To generate textual descriptions for each category in our dataset, $\mathcal{C}^D$, we use an LLM that provides detailed information about the animals without requiring expert input. 
The prompt used to extract these descriptions from the LLM is provided in Appendix \ref{supp: prompts}.
The LLM generates $M_c$ short descriptions for each class $c \in \mathcal{C}^D$. \method~assumes that this approach introduces more diverse information for representing each class. The generated descriptions are then processed by \method~using the text encoder.
To obtain the final embedding, the model computes the centroid of the resulting embeddings.
Specifically, let $\mathbf{P}^{(c)} \in \mathbb{R}^{M_c \times F}$ be the set of $M_c$ embeddings obtained from the LLM-generated descriptions for class $c$.
The {centroid} for each class $c$ is computed as:
\begin{equation}
\label{eq:centroid}
    \mathbf{t}^{(c)} = \frac{1}{M_c} \sum_{i=1}^{M_c} \mathbf{P}^{(c)}_{i},
\end{equation}
where $\mathbf{P}^{(c)}_{i}$ represents the $i$-th row of $\mathbf{P}^{(c)}$, corresponding to an individual textual description embedding.
Finally, the output of the text embedding component in \method~is a matrix:
\begin{equation}
    \mathbf{T} = [\mathbf{t}_1, \mathbf{t}_2, \dots, \mathbf{t}_{\vert \mathcal{C}^D \vert}]^\top \in \mathbb{R}^{\vert \mathcal{C}^D \vert \times F},
\end{equation}
which contains the final embeddings for all classes in $\mathcal{C}^D$.



\subsection{Image Encoder}
\label{sec:Image Embeddings}

\subsubsection{Pre-processing}
We employ an object detection model to process our camera trap datasets, aiming to extract image crops that contain relevant information for analysis.

\subsubsection{Image Embeddings}
\method~employs an image encoder to extract feature embeddings from cropped images.
The images are processed in mini-batches of size $B$, where each image is transformed into an embedding of dimension $F$ using the image encoder.
The output of this stage is a matrix:
\begin{equation}
    \mathbf{V} = [\mathbf{v}_1, \mathbf{v}_2, \dots, \mathbf{v}_B]^\top \in \mathbb{R}^{B \times F},
\end{equation}
where $\mathbf{v}_i $ represents the visual embedding of the $i $-th image in the mini-batch.  
This matrix is used as input for the subsequent stages of our framework, where text and image embeddings are aligned and contrasted.

\subsection{Image-text Encoder}
\label{sec:Image-text Embeddings}
\begin{figure}
    \centering
    \includegraphics[width=\columnwidth]{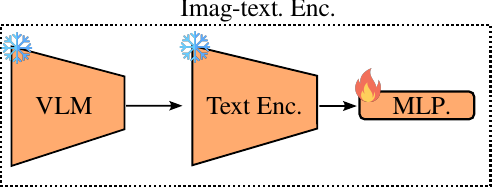}
    \caption{Detailed illustration of the image-text module, which consists of a VLM, a text encoder, and an MLP. This module processes input images and converts them into image-text embeddings.}
    \label{fig:image-text component}
\end{figure}

In the image-text branch of \method, we use the mini-batch of cropped images as input. To generate textual descriptions of the animals in these images, we utilize an image-text encoder, as illustrated in Figure~\ref{fig:image-text component}. This encoder consists of three main components: a VLM, a text encoder, and an MLP.
First, the VLM generates textual descriptions based on the input images, using a prompt similar to the one described in \citep{fabian2023knowledge} and provided in the Appendix \ref{supp: prompts}. 
Therefore, these textual descriptions are processed using the text encoder.
Finally, \method~applies an MLP to refine the extracted embeddings by introducing trainable parameters. As demonstrated in Section~\ref{sec:experiments_results}, incorporating trainable parameters improves the model’s performance. However, effective alignment between the image embeddings and the projected representations requires a dedicated similarity mechanism and a contrastive loss function, as detailed in Section~\ref{sec:embeddings_aligmnet} and Section~\ref{sec:contrative}.

The output of the image-text encoder of \method~is a matrix:
\begin{equation}
    \mathbf{L} = [\mathbf{l}_1, \mathbf{l}_2, \dots, \mathbf{l}_B]^\top \in \mathbb{R}^{B \times F},
\end{equation}
where $ \mathbf{l}_i $ represents the transformed embedding of the $ i $-th image description in the mini-batch.

\subsection{Similarity Mechanism}
\label{sec:embeddings_aligmnet}

The embeddings from the three types of modalities, text ($\mathbf{T}$), image ($\mathbf{V}$), and image-text ($\mathbf{L}$), are the inputs for the similarity method. The process incorporates two stages: i) similarity computation and ii) weighted integration. 
In the first stage, \method~computes the cosine similarities between text and image embeddings, as well as text and image-text embeddings. 
Specifically, let $\mathbf{W} \in \mathbb{R}^{B \times \vert \mathcal{C}^D \vert}$ be the matrix of cosine similarities between the text and image embeddings, computed as follows:
\begin{equation}
    \label{equation:similaties_W}
    \mathbf{W}_{ij} = \frac{\langle \mathbf{v}_i, \mathbf{t}_j \rangle}{\Vert \mathbf{v}_i\Vert \Vert \mathbf{t}_j \Vert}~\forall~1\leq i \leq B, 1\leq j \leq \vert \mathcal{C}^D \vert,
\end{equation} 
where $\mathbf{W}_{ij}$ represents the $(i,j)$ item of the matrix, $\langle \cdot, \cdot \rangle$ denotes inner product, and $\Vert \cdot \Vert$ is the $\ell_2$ norm of a vector.
Along the same process, \method~calculates the cosine similarities between the text and image-text embeddings as follows:
\begin{equation}
\label{equation:similaties_Q}
    \mathbf{Q}_{ij} = \frac{\langle \mathbf{l}_i, \mathbf{t}_j \rangle}{\Vert \mathbf{l}_i\Vert \Vert \mathbf{t}_j \Vert}~\forall~1\leq i \leq B, 1\leq j \leq \vert \mathcal{C}^D \vert,
\end{equation}
where $\mathbf{Q} \in \mathbb{R}^{B \times \vert \mathcal{C}^D \vert}$ is the matrix of cosine similarities between the text and image-text embeddings.

Both cosine similarities are combined using a weighted average between the matrices $\mathbf{W}$ and $\mathbf{Q}$, where the weights are controlled by the hyperparameter $\alpha \in [0,1]$.
Specifically, the output of the weighted integration is a matrix $\mathbf{S} \in \mathbb{R}^{B \times \vert \mathcal{C}^D \vert}$, defined as follows:
\begin{equation}
    \mathbf{S} = \alpha \mathbf{W} + (1- \alpha) \mathbf{Q}.
    \label{eqn:fusion}
\end{equation}
Since $\alpha \in [0,1]$, the resulting matrix $\mathbf{S}$ is a convex combination of $\mathbf{W}$ and $\mathbf{Q}$. As a result, each element $\mathbf{S}_{ij}$ in the matrix is also between $0$ and $1$.

\subsection{Contrastive Loss}
\label{sec:contrative}

We train our model using a contrastive loss function, $\mathcal{L}$, which takes the matrix $\mathbf{S}$ as input.
The loss function is calculated for each mini-batch as follows:
\begin{equation}
    \mathcal{L}(\mathbf{S}) = \frac{1}{B} \sum_{i=1}^B -\log \frac{\exp(\mathbf{S}_{ik}/\tau)}{\sum_{j=1}^{\vert \mathcal{C}^D \vert} \exp(\mathbf{S}_{ij}/\tau)},
    \label{eqn:loss}
\end{equation}
where $\tau$ is a temperature hyperparameter and $k$ is the index of the class in $\mathcal{C}^D$ of the $i$th image in the mini-batch.
This loss function aims to ensure that the embeddings corresponding to the same species category are brought closer together in the feature space. 

\section{Experiments and Results}
\label{sec:experiments_results}
In this section, we describe the datasets used in this work, the evaluation protocol, implementation details, results, and a discussion of \method. We compare our proposal with CLIP \citep{radford2021learning}, CLIP-Adapter \citep{gao2024clip}, Long-CLIP \citep{zhang2024long}, BioCLIP \citep{stevens2024bioclip}, WildCLIP \citep{gabeff2024wildclip}, and some adaptations of Long-CLIP and BioCLIP. Additionally, we conduct ablation studies to analyze the contribution of each component of \method, such as the image encoder, the image-text encoder, and LLM. We explore the effect of incorporating a template set to introduce task-specific information and evaluate different LLMs to assess their impact on performance. Finally, we investigate the sensitivity of \method~to the hyperparameter $\alpha$ in the similarity mechanism, and to the number of LLM-prompted sentences in the text encoder. All evaluations are reported using accuracy, and macro F1-score. \\

\subsection{Datasets}
\begin{figure}[t]
    \centering
    \includegraphics[width=\columnwidth]{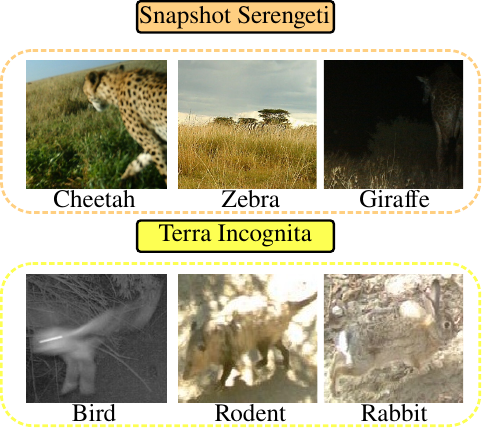}
    \caption{Cropped images from the Snapshot Serengeti and Terra Incognita datasets where we observe the geographical domain shift and the difference in classes (different taxonomic groups).}
    \label{fig:datasets}
\end{figure}

\begin{figure}[t]
    \centering
    \includegraphics[width=\columnwidth]{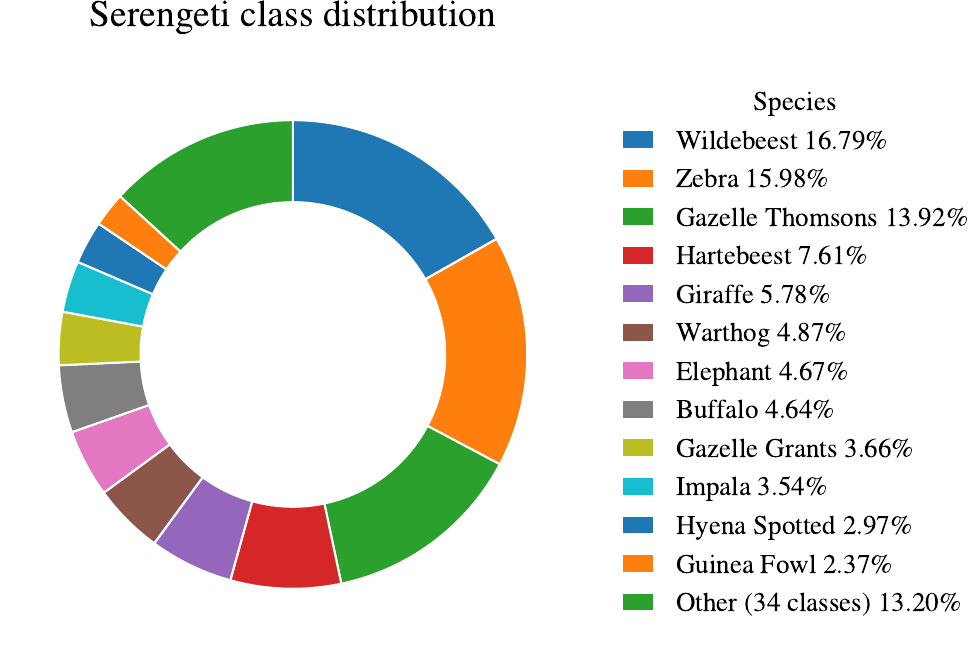}
    \caption{Class distribution of the Serengeti dataset.}
    \label{fig:serengeti_distribution}
\end{figure}

\begin{figure}[t]
    \centering
    \includegraphics[width=\columnwidth]{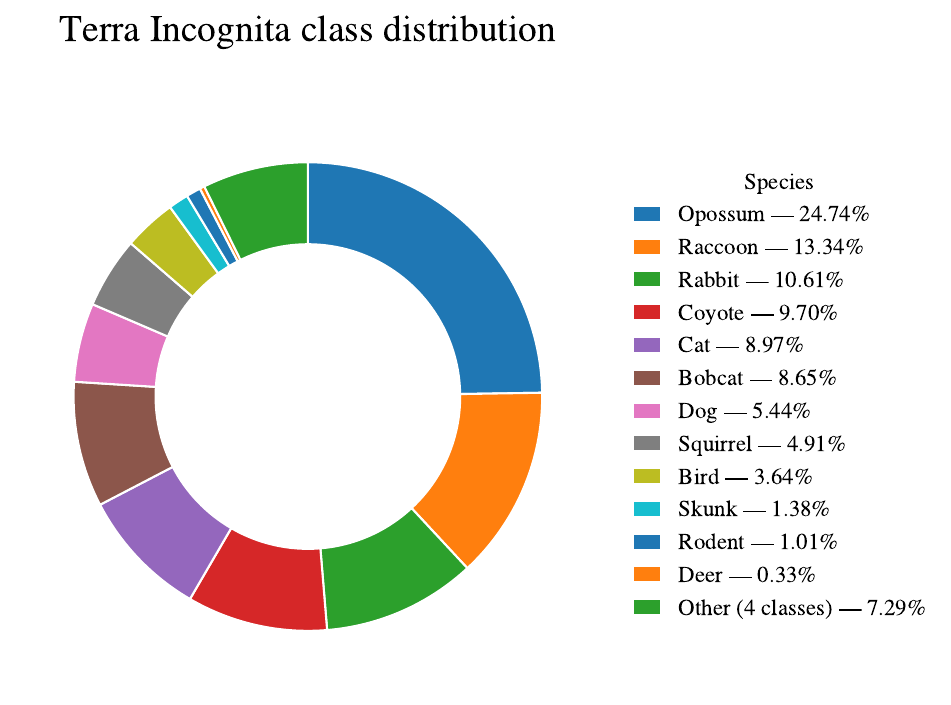}
    \caption{Class distribution of the Terra Incognita dataset.}
    \label{fig:terra_distribution}
\end{figure}

We evaluate \method~using two publicly available camera trap datasets from different geographical regions: Snapshot Serengeti \citep{dryad_5pt92}, collected in savanna environments in Africa using Scoutguard cameras, and Terra Incognita \citep{beery2018recognition}, collected in the southwest of the United States, where the predominant environment is semi-arid desert and pinyon–juniper woodland \citep{archer2008climate}. Information about the specific camera trap models used in the Terra Incognita dataset is not specified. This dataset presents several visual challenges, such as poor illumination (especially at night), motion blur due to low shutter speed, occlusions from vegetation or frame edges, and forced perspective when animals appear very close to the camera. Examples of cropped images from these datasets are shown in Figure~\ref{fig:datasets} and their class distributions are shown in Figure~\ref{fig:serengeti_distribution} and Figure~\ref{fig:terra_distribution}. 

\begin{itemize}[1.]
    \item \textit{Snapshot Serengeti} \citep{dryad_5pt92}.  
    We use the version of the Snapshot Serengeti dataset adopted in WildCLIP \citep{gabeff2024wildclip}, which consists of 46 classes.  
    This dataset version contains $380 \times 380$ pixel image crops, generated by the MegaDetector model from the Snapshot Serengeti project, using a confidence threshold above $0.7$.  
    Only images containing single animals were selected. The dataset includes a total of $340,972$ images, with $230,971$ for training, $24,059$ for validation, and $85,942$ for testing.  

    \item \textit{Terra Incognita} \citep{beery2018recognition}.  
    This dataset consists of 16 classes and introduces two testing groups: Cis-locations and Trans-locations.  
    Cis-locations contain images similar to the training data, while Trans-locations feature images from different environments.  
    These partitions were originally designed to assess the robustness of computer vision models in an in-domain evaluation setting.  
    We filter the images in this dataset using the MegaDetector model from the PyTorch-Wildlife library \citep{hernandez2024pytorchwildlife}.  
    The dataset contains a total of $45,912$ images, distributed as follows: $12,313$ for training, $1,932$ for Cis-validation, $1,501$ for Trans-validation, $13,052$ for Cis-test, and $17,114$ for Trans-test.  
\end{itemize}

\subsection{Evaluation Protocol}
We conduct two experiments to assess the performance of our model in comparison to state-of-the-art (SOTA) methods.  
In the first experiment, we use the Snapshot Serengeti dataset for training and validation and the Terra Incognita dataset for testing. This cross-dataset setup allows us to evaluate the model's performance under geographical domain shift, i.e., geographical out-of-domain evaluation.
Formally, we define $\mathcal{D}$ as the Snapshot Serengeti dataset and $\mathcal{S}$ as the Terra Incognita dataset.  
Snapshot Serengeti was collected in various protected areas in Africa, whereas Terra Incognita originates from the southwest of the United States. 
This experimental setup introduces two key challenges:  
i) a distribution shift between the datasets $\mathcal{D}$ and $\mathcal{S}$ (geographical domain shift), and  
ii) a discrepancy in the set of classes, where $\mathcal{C}^D \neq \mathcal{C}^S$.  
These challenges are illustrated in Figure \ref{fig:datasets}.  
Due to the difference in class sets, closed-set SOTA methods cannot be used for comparison.  
We report accuracy results on the Cis-Test and Trans-Test subsets of Terra Incognita.  

In the second experiment, we modify our problem definition from Section~\ref{sec:Method} and use the same dataset for training, validation, and testing to evaluate the model without the challenges introduced by the geographical domain shift and novel classes, i.e., under in-domain evaluation. 
Specifically, we train and evaluate the model on either the Snapshot Serengeti or the Terra Incognita datasets.  
This approach allows us to assess the model's accuracy and robustness within a consistent domain.

\begin{table*}[t]
\centering
\caption{Zero-shot performance results of \method~and other foundation models on the Terra Incognita dataset (out-of-domain evaluation). All methods are trained on data different from the test dataset. The best method is highlighted in \textbf{bold}, and the second-best is \underline{underlined}. Results are reported in accuracy (\%) and macro F1 score (F1-M)}.
\label{tab:zero-shot-results}
\begin{tabular}{lccccccc}
\toprule
\textbf{Model}  & \textbf{Training}  & \textbf{Cis-Test}  & \textbf{F1-M}  &\textbf{ \textbf{Trans-Test} } & \textbf{F1-M} \\ 
\midrule 
CLIP  & OpenAI data  & $39.14$ & $0.39 $& $34.67$ & $0.32$  \\ 
Long-CLIP  & ShareGPT4V & $42.41$ & $0.41$ & $37.55$ & $0.34$  \\ 
BioCLIP  & TREEOFLIFE-10M  & $21.12$ & $0.20$ & $14.53$ & $0.15$ \\
\hdashline
CLIP Adapter & Snapshot Serengeti  & $27.45_{\footnotesize{\pm 2.84}}$ & $0.25_{\footnotesize{\pm 0.03}}$ & $16.17_{\footnotesize{\pm 3.37}}$ & $0.18_{\footnotesize{\pm 0.03}}$ \\ 
Long-CLIP Adapter & Snapshot Serengeti & $30.40_{\footnotesize{\pm 1.58}}$ & $0.28_{\footnotesize{\pm 0.02}}$ & $18.73_{\footnotesize{\pm 1.54}}$ & $0.19_{\footnotesize{\pm 0.02}}$ \\ 
BioCLIP Adapter & Snapshot Serengeti  & $14.21_{\footnotesize{\pm 3.30}}$ & $0.12_{\footnotesize{\pm 0.03}}$ & $8.59_{\footnotesize{\pm 3.02}}$  & $0.08_{\footnotesize{\pm 0.01}}$ \\
WildCLIP    & Snapshot Serengeti  & $41.62_{\footnotesize{\pm 0.40}}$ & $0.39_{\footnotesize{\pm 0.01}}$ & $37.52_{\footnotesize{\pm 0.42}}$ & $0.31_{\footnotesize{\pm 0.01}}$ \\ 
WildCLIP-LwF  & Snapshot Serengeti  & $\underline{43.67}_{\footnotesize{\pm 0.12}}$ & $\underline{0.40}_{\footnotesize{\pm 0.01}}$ &$\textbf{40.17}_{\footnotesize{\pm 0.05}}$  & $\underline{0.34}_{\footnotesize{\pm 0.01}}$ \\ 
\hdashline
\textbf{\method~(ours)} & Snapshot Serengeti & $\textbf{50,06}_{\footnotesize{\pm 2.39}}$ & $\textbf{0.43}_{\footnotesize{\pm 0.01}}$ & $\underline{39.96}_{\footnotesize{\pm 1.89}}$  & $\textbf{0.36}_{\footnotesize{\pm 0.01}}$ \\
\bottomrule
\end{tabular}
\end{table*}

\subsubsection{Implementation Details}
For pre-processing, we employ the MegaDetector model \citep{beery2019efficient}. \method~is implemented using the 3.5 version of ChatGPT for the LLM, the LongCLIP-B version of Long-CLIP for the text and image encoder, and the 1.5-7B version of LLaVA for the VLM.
For training our model in both experiments, we set $\tau = 0.1$.  
The MLP architecture for the first experiment consists of a single hidden layer with a dimension of $793$ and employs the Rectified Linear Unit (ReLU) as the activation function. Additionally, a skip connection is implemented between the input and output layers.  
Additionally, we train the model for $30$ epochs and set $\alpha = 0.5$.  
Optimization is performed using the Stochastic Gradient Descent (SGD) algorithm with a learning rate of $0.09$, a momentum of $0.80$, and a batch size of $128$.  
We perform our experiments on GPUs Tesla P100-PCIE-16GB.

In the second experiment, we fine-tune \method~by unfreezing the image encoder and adjusting key hyperparameters, including the $\alpha = 0.6$. 
Specifically, we use a batch size of $256$ and train the model for $57$ epochs using SGD with a momentum of $0.82$ and a learning rate of $1\mathrm{e}{-3}$.  
For the Snapshot Serengeti dataset, we apply an MLP with a single hidden layer with $256$ dimensions.  
For the Terra Incognita dataset, we set an MLP with a single hidden layer of $733$ dimensions.  
Early stopping is applied with a patience of $5$ epochs.  

To optimize the hyperparameters, we use a random search. Furthermore, we evaluate the best hyperparameter combination using $100$ different random seeds for the first experiment and $3$ random seeds for the second experiment.
For details on the search space and additional information, please refer to Appendix~\ref{supp: hyperparameter search}.

\subsection{Quantitative Results}
\subsubsection{Comparison with the SOTA in Out-of-domain Evaluation}
Table \ref{tab:zero-shot-results} presents the zero-shot performance evaluation of various models trained on datasets such as ShareGPT4V and Snapshot Serengeti and evaluated on the Cis-Test and Trans-Test sets of the Terra Incognita dataset. Models such as CLIP, BioCLIP, and Long-CLIP (first three rows) are reported without their standard deviation, as we used the pre-trained models provided by the authors of each model.

For the remaining cases, we report the results obtained by training the model with $100$ different random seeds. The results indicate that CLIP, trained on OpenAI data, achieves a Cis-Test accuracy of $39.14\%$ and a Trans-Test accuracy of $34.67\%$, while Long-CLIP, trained on ShareGPT4V, improves these metrics to $42.41\%$ and $37.55\%$, respectively. BioCLIP, despite being trained on TREEOFLIFE-10M, struggles to generalize to the Terra Incognita dataset, achieving only $21.12\%$ on Cis-Test and $14.53\%$ on Trans-Test. 
Based on CLIP Adapter \citep{gao2024clip}, we evaluate this strategy and extend it to Long-CLIP and BioCLIP. The results for these adapter-based models show lower performance compared to models without adapters. Specifically, the original CLIP Adapter achieves an accuracy of $27.45\%$ in Cis-Test. The Long-CLIP Adapter and BioCLIP Adapter variants obtain $30.4\%$ and $14.21\%$ accuracy in Cis-Test, while their Trans-Test accuracies are $16.17\%$, $18.73\%$, and $8.59\%$, respectively. 
Overall, these results suggest that adapter-based models are highly domain-specific and further widen the gap between domains compared to the original architectures.

\begin{table}[t]
\caption{Performance comparison in Snapshot Serengeti (in-domain evaluation). All models use the ViT-B/16 backbone. The best method is highlighted in \textbf{bold}, and the second-best is \underline{underlined}. Results are reported in accuracy (\%).}
\centering
\begin{tabular}{lcc}
\toprule
\textbf{Model} & \textbf{Loss Function} & \textbf{Test} \\ \midrule
Linear Probe CLIP & Cross-entropy & $\underline{84.84}_{\footnotesize{\pm 0.42}}$ \\
\hdashline
CLIP Adapter & Contrastive & $84.77_{\footnotesize{\pm 0.26}}$ \\ 
Long-CLIP Adapter & Contrastive & $83.97_{\footnotesize{\pm 1.03}}$ \\ 
BioCLIP Adapter & Contrastive & $80.47_{\footnotesize{\pm 0.89}}$  \\
WildCLIP    & Contrastive & $68.73_{\footnotesize{\pm 0.28}}$   \\ 
WildCLIP-LwF  & Contrastive & $69.53_{\footnotesize{\pm 0.02}}$   \\ 
\hdashline
\textbf{\method~(ours)} & Contrastive  & $\textbf{90.74}_{\footnotesize{\pm 0,05}}$ \\ \bottomrule 
\end{tabular}
\label{tab:performance serengeti}
\end{table}

WildCLIP and its Learning without Forgetting (LwF) variant, both trained on Snapshot Serengeti, show better performance, achieving $41.62\%$ and $43.67\%$ on Cis-Test, and $37.52\%$ and $40.17\%$ on Trans-Test, respectively. Our proposed method, \method, outperforms all previously evaluated models on Cis-Test, achieving $50.06\%$ accuracy, and obtains the second-best accuracy on Trans-Test with $39.96\%$, demonstrating its effectiveness in geographical out-of-domain evaluation. Furthermore, the macro F1 scores highlight that \method~is less biased toward majority classes (see Fig.~\ref{fig:terra_distribution} for the class distribution of Terra Incognita), this behavior is observed in both test sets (Cis-Test and Trans-Test). \method~is the only model that improves the macro F1 score relative to its base model (LongCLIP in our case), increasing it from 0.41 to 0.45 on Cis-Test and from 0.34 to 0.37 on Trans-Test. These results highlight that \method~not only generalizes well across domains, but also improves per-class performance despite being trained on a highly imbalanced dataset such as Snapshot Serengeti (see Fig.~\ref{fig:serengeti_distribution}).

In addition, the standard deviation of our proposal is comparable to that of SOTA models such as CLIP Adapter, Long-CLIP Adapter, and BioCLIP Adapter.
These results highlight the advancements of \method~in learning geographical domain-invariant representations and improving open-set recognition. By leveraging additional semantic information to represent the input, \method~surpasses previous SOTA models in handling geographical domain shifts and recognizing unseen classes.

\subsubsection{Trainable Parameters and Computational Cost }

In our comparison of trainable parameters, we exclude the zero-shot models CLIP, Long-CLIP, and BioCLIP, since these models are not trained as part of this work. \method~has a total of 813,339 trainable parameters. This is significantly lower than the WildCLIP and WildCLIP-LwF models, which have approximately 86 million trainable parameters. This difference is because those models unfreeze the CLIP image encoder during training. Nevertheless, \method~achieves best performance in out-of-domain evaluation on the Cis-Test set, with an accuracy of 50.06\% and ranks second on the Trans-Test set with an accuracy of 39.96\% (see Table \ref{tab:zero-shot-results}). On the other hand, the adapter-based models (CLIP Adapter, Long-CLIP Adapter, and BioCLIP Adapter) have the smallest number of trainable parameters, with 262,914. However, these adapter-based models achieved the worst performance, as shown in Table~\ref{tab:zero-shot-results}. Finally, it is clear that \method~offers a good trade-off between model complexity and performance.

For computational cost, training the full model of \method~on the Snapshot Serengeti dataset takes around 30 minutes using a single Tesla P100-PCIE-16GB GPU. For comparison, training the standard version of WildCLIP \citep{gabeff2024wildclip} in the same dataset on A100-PCIE-40GB takes 13.4 hours. For that reason, it is clear that our proposal offers a lightweight alternative for addressing geographical domain shift, instead of training a large model such as WildCLIP.


\subsubsection{In-domain Performance Comparison in the Snapshot Serengeti Dataset}
\label{sec:in-domain serengeti}
Table \ref{tab:performance serengeti} shows a comparison of multiple models trained in the Snapshot Serengeti dataset.
The Linear Probe CLIP model, trained with a cross-entropy loss, achieves a test accuracy of $84.84\%$.
Despite this performance, it lacks open vocabulary capabilities due to its supervised training approach.
Among the adapter-based models, the CLIP Adapter, Long-CLIP Adapter, and BioCLIP Adapter achieve test accuracies of $84.77\%$, $83.97\%$, and $80.47\%$, respectively. These results suggest that adapter-based methods are a good option for in-domain evaluation. 
WildCLIP achieves a test accuracy of $68.73\%$, demonstrating moderate performance but performing worse than both adapter-based models and our method. The WildCLIP-LwF variant improves this performance, reaching $69.53\%$. The LwF strategy contributes positively to retaining learned information, but its improvement remains limited compared to other models like CLIP Adapter and \method. 
The proposed method, \method, achieves the highest test accuracy of $90.74\%$, outperforming all previously evaluated models while maintaining a low standard deviation. 

\begin{table}[t]
\caption{Performance comparison in Terra Incognita (in-domain evaluation). All models use the ViT-B/16 backbone. The best method is highlighted in \textbf{bold}, and the second-best is \underline{underlined}. Results are reported in accuracy (\%).}
\centering
\begin{tabular}{lcc}
\toprule
\textbf{Model} & \textbf{Cis-Test} & \textbf{Trans-Test} \\ \midrule
Linear Probe CLIP  & $78.09_{\footnotesize{\pm 0.34}}$ & $67.23_{\footnotesize{\pm 2.31}}$ \\
\hdashline
CLIP Adapter & $79.45_{\footnotesize{\pm 1.21}}$ & $69.10_{\footnotesize{\pm 2.70}}$ \\ 
Long-CLIP Adapter & $79.11_{\footnotesize{\pm 0.47}}$ & $69.15_{\footnotesize{\pm 1.33}}$ \\ 
BioCLIP Adapter & $77.09_{\footnotesize{\pm 1.19}}$ & $63.45_{\footnotesize{\pm 0.85}}$  \\
\textbf{WildCLIP}    & $\textbf{91.70}_{\footnotesize{\pm 0.21}}$ & $\textbf{84.47}_{\footnotesize{\pm 0.24}}$   \\ 
WildCLIP-LwF  & $\underline{88.93}_{\footnotesize{\pm 0.02}}$ & $\underline{82.91}_{\footnotesize{\pm 0.01}}$   \\ 
\hdashline
\method~(ours) & $84.10_{\footnotesize{\pm 3.03}}$  & $75.80_{\footnotesize{\pm 5.38}}$ \\ \bottomrule 
\end{tabular}
\label{tab:performance Terra}
\end{table}

\subsubsection{In-domain Performance Comparison in the Terra Incognita Dataset}
Similar to Section \ref{sec:in-domain serengeti}, Table \ref{tab:performance Terra} presents a comparison of different models in the geographical in-domain evaluation. All models in this comparison are trained and evaluated on the Terra Incognita dataset.

The Linear Probe CLIP model achieves a Cis-Test accuracy of $78.09\%$ and a Trans-Test accuracy of $67.23\%$. 
Among the adapter-based methods, the CLIP Adapter, Long-CLIP Adapter, and BioCLIP Adapter achieve Cis-Test accuracies of $79.45\%$, $79.11\%$, and $77.09\%$, respectively, while their Trans-Test accuracies are $69.10\%$, $69.15\%$, and $63.45\%$. These results indicate that adapter-based models provide competitive performance in in-domain evaluation.

WildCLIP achieves a Cis-Test accuracy of $91.70\%$ and a Trans-Test accuracy of $84.47\%$, demonstrating the highest performance among all evaluated models. The LwF variant of WildCLIP slightly underperforms its base version, reaching a Cis-Test accuracy of $88.93\%$ and a Trans-Test accuracy of $82.91\%$. 

The proposed method achieves a Cis-Test accuracy of $84.10\%$, outperforming the adapter-based models and the Linear Probe CLIP. In the Trans-Test scenario, \method~achieves $75.80\%$, showing competitive performance but lower than WildCLIP and WildCLIP-LwF. 
However, we observe a significant increase in the standard deviation due to the smaller size of the Terra Incognita dataset, which limits the robustness when evaluated with only three random seeds. This effect is further amplified by unfreezing the image encoder and fine-tuning it in this experiment. It is well known in the domain generalization literature that there is an inherent trade-off between improving accuracy on a specific domain and achieving effective domain alignment for generalization across domains \citep{nguyen2022trade,yu2024rethinking}. To address this limitation, few-shot adaptation and training-free methods could be explored to reduce the dependence on large datasets \citep{zhang2022tip,zanella2024low,bendou2025proker}. In future work, we will  explore these approaches, as their investigation falls beyond the scope of the current study.

\subsection{Ablation Studies}
\label{sec:ablations}
We conduct several ablation studies on (i) the impact of incorporating a template set to introduce task-specific information; (ii) the impact of the image encoder, the image-text encoder, and the LLM; and (iii) the influence of different LLMs on performance. 
This analysis is conducted for the out-of-domain evaluation.

\subsubsection{Evaluating the Incorporation of a Template Set}
\label{sec:templates}
We evaluate the impact of using a predefined template set to describe the dataset categories, following the original configuration used in CLIP \citep{radford2021learning}. This approach aims to incorporate task-specific information captured by the template set and serves as a potential alternative to LLM-generated descriptions in scenarios where their use is impractical or computationally expensive.
The template set adds context to the descriptions by specifying that the images were captured by camera traps. For example, a template might describe a category as: ``A photo captured by a camera trap of a \{ \}''.
For a detailed list of the template set, refer to Appendix~\ref{supp: templates}.

\begin{figure}[t]
    \centering
     \includegraphics[width=\columnwidth]{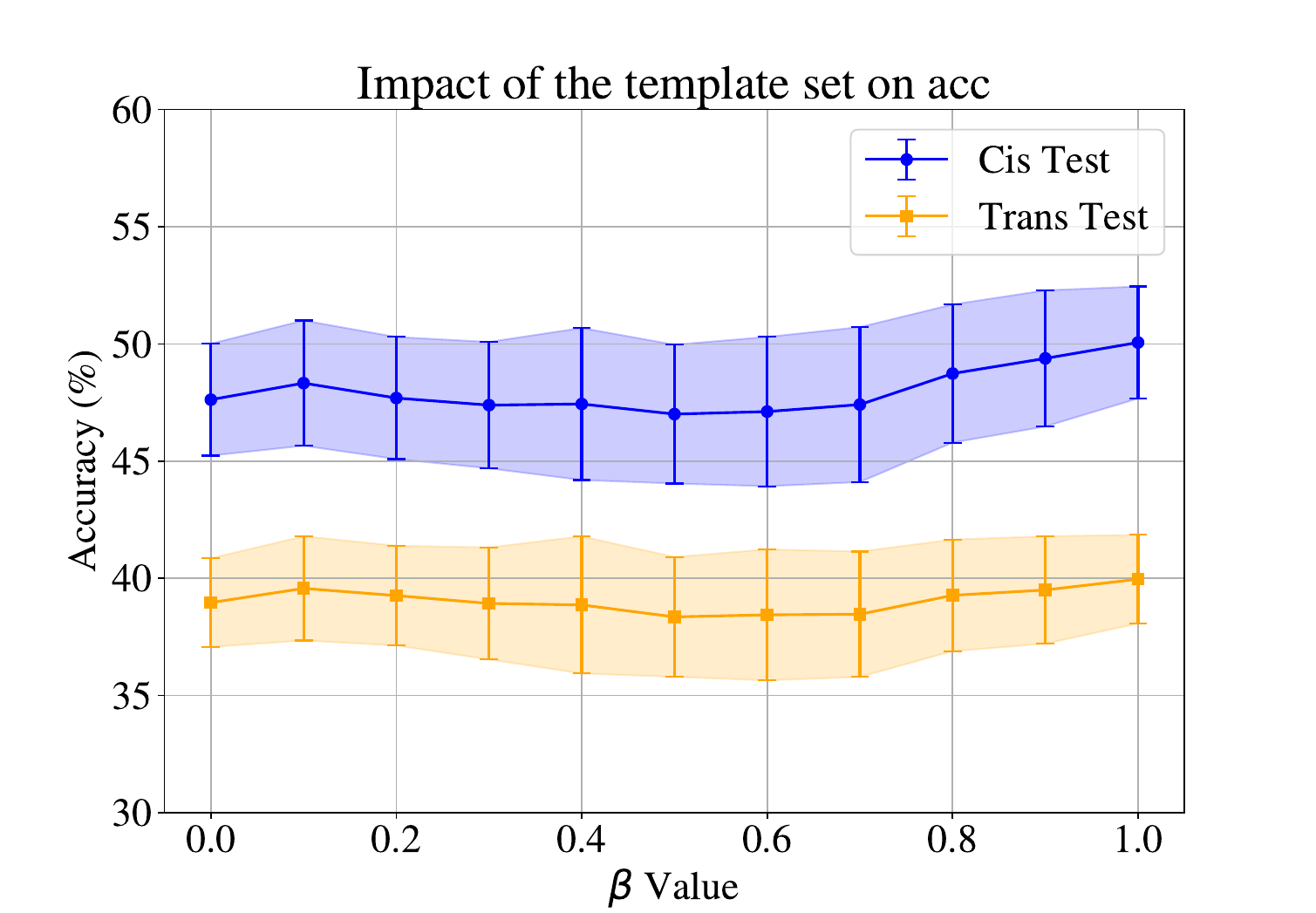}
    \caption{Evaluation of the template set contribution to the \method~in out-of-domain performance.}
    \label{fig:analysis_beta}
\end{figure}

To integrate template-based descriptions with LLM descriptions, we introduce a hyperparameter $\beta$, which controls the contribution of each source of knowledge to the final text embedding for each class. For both the template set and LLM-generated descriptions, we compute the centroid as defined in equation \eqref{eq:centroid}. 
The final text embedding is obtained as a weighted combination of these two centroids $\mathbf{t}_c = (1 - \beta) \mathbf{m}_1^{(c)} + \beta \mathbf{m}_2^{(c)}$, 
where $\mathbf{m}_1^{(c)}$ represents the centroid of the template-based descriptions, and $\mathbf{m}_2^{(c)}$ represents the centroid of the LLM-based descriptions. The parameter $\beta \in [0,1]$ determines the relative influence of each source.  


Figure \ref{fig:analysis_beta} presents the impact of the template set on model performance by varying the hyperparameter $\beta$. Figure \ref{fig:analysis_beta} shows that the best results are achieved when $\beta = 1$, which corresponds to excluding the template set from the text embedding calculation.

\begin{table}[t]
\caption{Ablation studies for performance variations for different design choices of \method. All models are trained on Snapshot Serengeti and evaluated on Terra Incognita (out-of-domain evaluation). The best combination is highlighted in \textbf{bold}, and the second-best is \underline{underlined}. Results are reported in accuracy (\%).}
\centering
\setlength{\tabcolsep}{2pt}
\begin{tabular}{cccccc}
\toprule
\textbf{Img Enc} & \textbf{Img-txt Enc} & \textbf{LLM} & \textbf{Cis-Test} & \textbf{Trans-Test} \\ 
\midrule
\cmark & \xmark & \xmark & $46.67$ & $37.18$  \\
\cmark & \xmark & \cmark & $40.32$ & $\underline{39.11}$  \\ 
\hdashline
\xmark & \cmark & \xmark & $28.14_{\footnotesize{\pm 2.53}}$ & $21.72_{\footnotesize{\pm 1.79}}$  \\
\xmark & \cmark & \cmark & $33.57_{\footnotesize{\pm 2.81}}$ & $26.66_{\footnotesize{\pm 2.96}}$  \\
\hdashline
\cmark & \cmark & \xmark & $\underline{47.62}_{\footnotesize{\pm 2.39}}$ & $38.96_{\footnotesize{\pm 1.89}}$  \\ 
\cmark & \cmark & \cmark &  $\textbf{50.06}_{\footnotesize{\pm 2.39}}$ & $\textbf{39.96}_{\footnotesize{\pm 1.89}}$ \\
\bottomrule
\end{tabular}
\label{tab:ablation_study1}
\end{table}

\subsubsection{Image Encoder, Image-text Encoder, and LLM}
We evaluate the performance of \method~by removing the image encoder, the image-text encoder, and the textual descriptions generated by the LLM. Table \ref{tab:ablation_study1} presents the geographical out-of-domain evaluation results for Cis-Test and Trans-Test under different design choices in the model.
Our findings show that the lowest performance occurs when both the image encoder and LLM-generated descriptions are removed (third row in Table \ref{tab:ablation_study1}). This setting is equivalent to setting $\alpha=0$ in \eqref{eqn:fusion} and training the model using the camera trap template set introduced in Section \ref{sec:templates}.
Under this configuration, the model achieves accuracy scores of $28.14\%$ in the Cis-Test and $21.72\%$ in the Trans-Test. Similarly, removing only the image encoder also results in poor performance (fourth row in Table \ref{tab:ablation_study1}). These results highlight the crucial role of the image encoder in \method, indicating that the image-text encoder alone cannot fully replace it.

\begin{table}[t]
\caption{Performance comparison with different LLMs. The best combination is highlighted in \textbf{bold}, and the second-best is \underline{underlined}. Results are reported in accuracy (\%).}
\centering
\begin{tabular}{lcc}
\toprule
\textbf{LLM} & \textbf{Cis-Test Acc(\%)} & \textbf{Trans-Test} \\ \midrule
LLAMA  & $28.82_{\footnotesize{\pm 0.84}}$ & $21.31_{\footnotesize{\pm 0.60}}$ \\
Qwen  & $\underline{29.70}_{\footnotesize{\pm 0.44}}$ & $\underline{22.88}_{\footnotesize{\pm 0.55}}$ \\
Phi  & $29.47_{\footnotesize{\pm 0.62}}$ & $19.96_{\footnotesize{\pm 0.66}}$ \\
\hdashline
ChatGPT & $\textbf{50.06}_{\footnotesize{\pm 2.39}}$ & $\textbf{39.96}_{\footnotesize{\pm 1.89}}$ \\ \bottomrule 
\end{tabular}
\label{tab:performance LLMs}
\end{table}

When the image encoder is included, the model shows a significant improvement, achieving accuracies of $46.67\%$ in the Cis-Test and $37.18\%$ in the Trans-Test (first row in Table \ref{tab:ablation_study1}).
Similar to the configuration in the first row of Table \ref{tab:ablation_study1}, adding LLM-generated descriptions improves performance in the Trans-Test, increasing accuracy to $39.11\%$. However, in the Cis-Test, the accuracy decreases to $40.32\%$ (second row in Table \ref{tab:ablation_study1}). These results suggest that while textual descriptions can be beneficial, their effectiveness depends on the test dataset and the type of text information used. This suggests that using only the image encoder is insufficient to capture geographically invariant features.
Standard deviations are not reported in the first two rows in Table \ref{tab:ablation_study1}, as we exclusively used the pre-trained model provided in Long-CLIP \citep{zhang2024long}.

When both image and image-text encoders are included, the model provides more robust results across both test sets (fifth and sixth row in Table \ref{tab:ablation_study1}). In particular, incorporating the image encoder, image-text encoder, and LLM-generated descriptions leads to the highest accuracy, with $50.06\%$ in the Cis-Test and $39.96\%$ in the Trans-Test.
This highlights the importance of integrating image and image-text embeddings more effectively to capture relationships between images and their categorical descriptions, helping to construct invariant representations against geographical domain shifts.

\subsubsection{Evaluating different LLMs}
Table \ref{tab:performance LLMs} shows a comparison of \method~when trained with descriptions generated by different LLMs, including LLAMA \citep{touvron2023llama}, Qwen \citep{yang2024qwen2}, Phi \citep{abdin2024phi}, and ChatGPT. The key difference among these models is the quality of the generated descriptions for each category. 


We observe that \method~achieves an accuracy of $28.82\%$ in the Cis-Test and $21.31\%$ in the Trans-Test when trained with descriptions from LLAMA. Similarly, poor results are obtained with Qwen and Phi.
In contrast, when using ChatGPT-generated descriptions, the model reaches $50.06\%$ in the Cis-Test and $39.96\%$ in the Trans-Test. 
This suggests that the descriptions generated by ChatGPT are significantly better than those produced by the other models in this specific task.
These results highlight the importance of generating high-quality descriptions that provide the model with relevant information about each category.
More informative descriptions enhance the model’s ability to distinguish between different categories, which in turn results in better performance across both test scenarios.


\subsection{Sensitivity to the Hyperparameter $\alpha$}

\begin{figure}[t]
    \centering
    \includegraphics[width=\columnwidth]{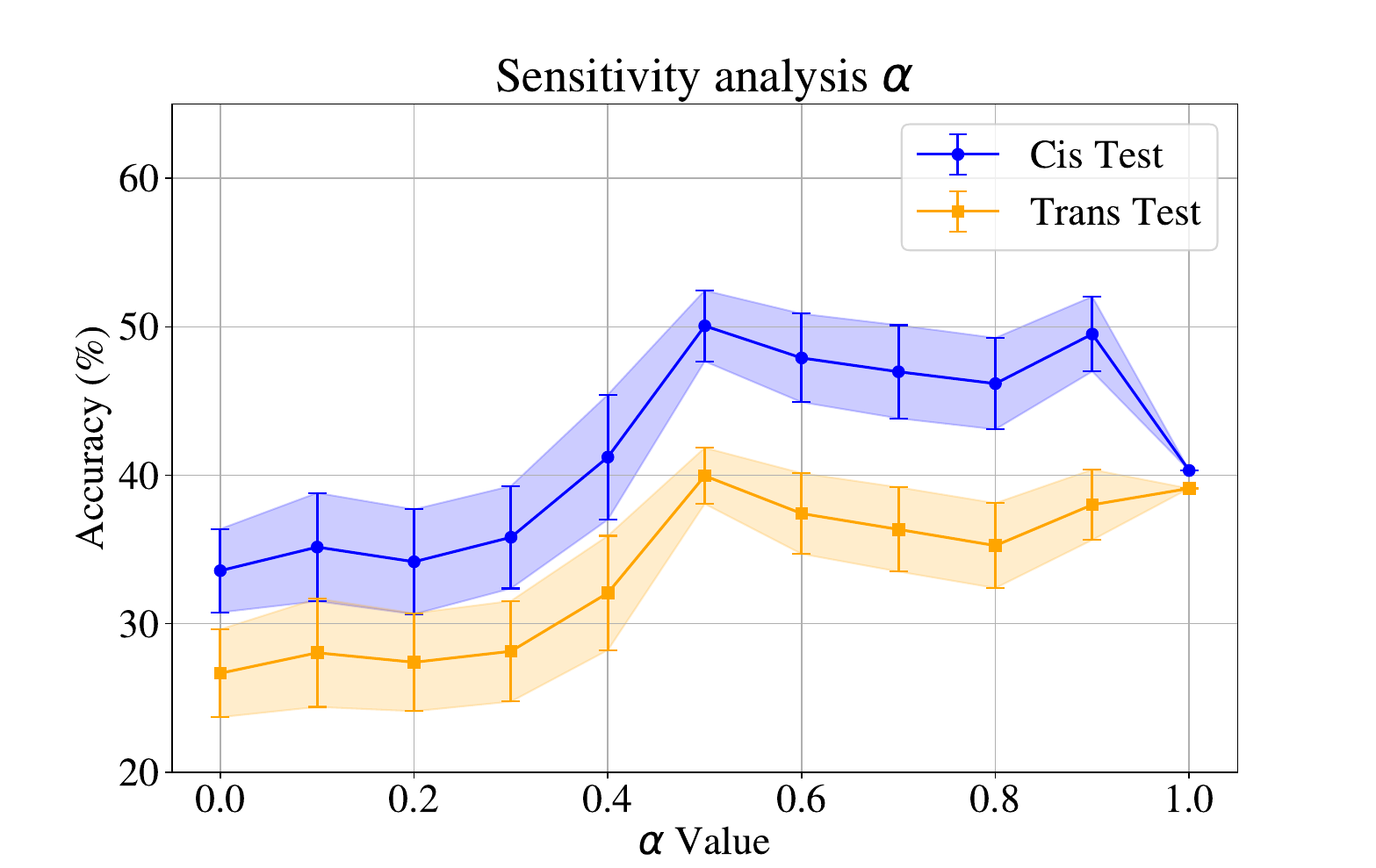}
    \caption{Sensibility analysis of the hyperparameter $\alpha$ of \method~in the Terra Incognita dataset for out-of-domain evaluation.}
    \label{fig:analysis_alpha}
\end{figure}

Figure~\ref{fig:analysis_alpha} illustrates how variations in the parameter $\alpha$ between $0$ and $1$ in \eqref{eqn:fusion} affect the performance of \method~for Cis-Test and Trans-Test in Terra Incognita for out-of-domain evaluation. 
We observe that the information from both matrices, $\mathbf{Q}$ and $\mathbf{W}$, is complementary. Figure~\ref{fig:analysis_alpha} shows that the optimal value of $\alpha$ is $0.5$, indicating that giving nearly equal importance to both matrices results in the highest accuracy. 
When $\alpha$ deviates from this optimal value, the accuracy declines, suggesting that overemphasizing either matrix leads to a loss of useful information for classification. This trend is consistent across both evaluation sets, demonstrating the robustness of the model when incorporating information from both matrices.

\begin{figure}[t]
    \centering
    \includegraphics[width=0.88\columnwidth]{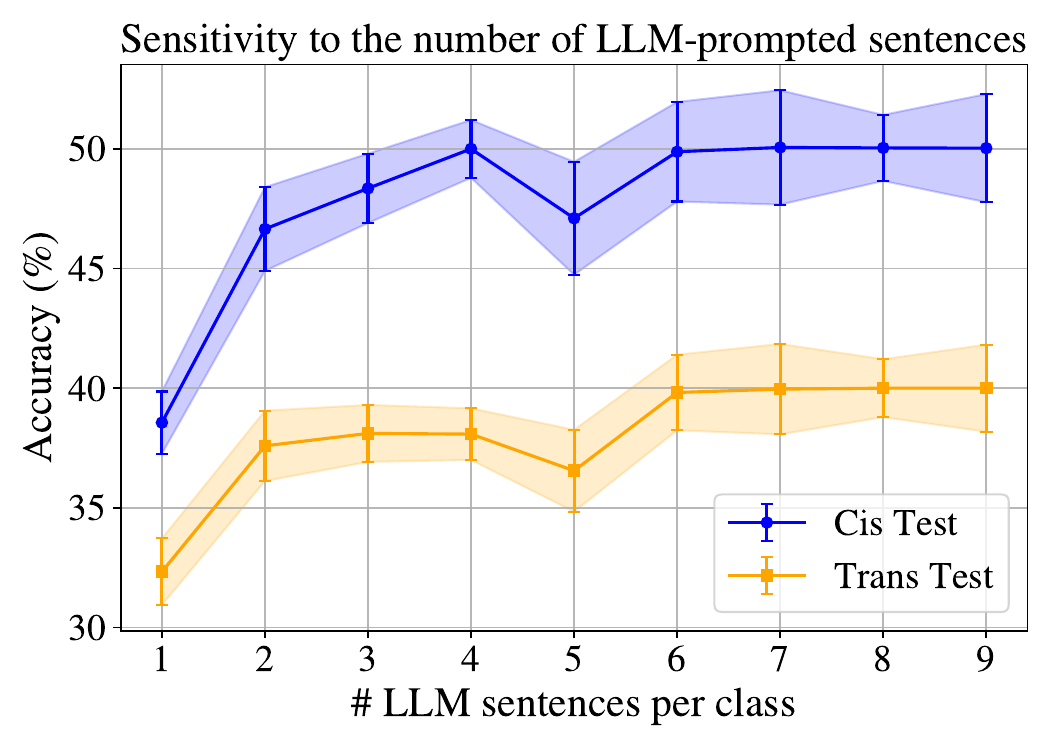}
    \caption{Evaluation of the sensitivity of \method~to the number of LLM-prompted sentences per class.}

    \label{fig:analysis_llm_sents}
\end{figure}

\subsection{Sensitivity to the number of LLM-prompted sentences}

The Figure~\ref{fig:analysis_llm_sents} present the performance of \method~when we vary $M_c$, the number of short description per class $c$ in $\mathcal{C}^D$.  We observe a consistent improvement in accuracy with more descriptions, especially on the Cis-Test split. These results suggest that \method~benefits from a larger number of short descriptions that are diverse and clearly describe each class.

\subsection{Limitations}

Although FMs have shown promising results in recognizing animal species in camera trap images, there is still a difference between their performance on out-of-domain (Table \ref{tab:zero-shot-results}) and in-domain (Table \ref{tab:performance Terra}) geographical evaluation. 

\begin{figure}[t]
    \centering
    \includegraphics[width=0.48\textwidth]{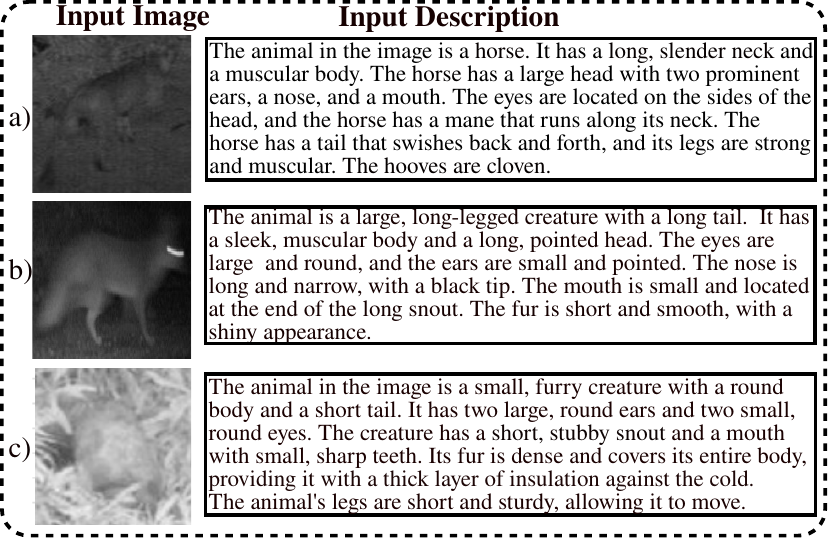}
    \caption{Failure cases of the \method~model in camera trap image classification. \textbf{Case a:} The VLM generates an incorrect description containing details that do not match the input image. \textbf{Case b:} A blurry image leads to a vague and uninformative description. \textbf{Case c:} When the input image is highly unclear, the VLM produces a random and unrelated description.}
    \label{fig:failure_cases}
\end{figure}


Figure~\ref{fig:failure_cases} illustrates the typical failure cases of \method.
These misclassifications arise mostly when descriptions are inaccurate or ambiguous. To illustrate the model's sensitivity to input descriptions, Figure~\ref{fig:failure_cases} presents three examples:  




\begin{itemize}[1.]
    \item In \textbf{case a}, the VLM generates hallucinated details that are not present in the image. For example, it identifies the animal as a horse, even though the input image does not match this description. This results in a completely incorrect prediction.

    \item In \textbf{case b}, the input image is blurry and difficult to interpret. Consequently, the VLM produces a vague description with minimal useful information. Due to the lack of context, the model fails to make an accurate prediction.

    \item In \textbf{case c}, the image is so unclear that the VLM generates a description unrelated to the input. This random description further misleads the model, leading to an incorrect prediction.
\end{itemize}

There are diverse Uncertainty Estimation (UE) strategies for handling hallucinated or low-quality captions \citep{li2025mitigating}. Neighborhood consistency can be used to identify likely unreliable model responses \citep{khan2024consistency}. The method proposed in \citep{khan2024consistency} introduces $n$ variations of an input prompt to the model (e.g., a VLM) and expects the evaluated model to produce consistent outputs for all $n$ cases, highlighting whether the model's deterministic embeddings can effectively handle semantic variability. Another solution, more on the hidden embeddings of the model, is proposed in \citep{mushtaq2025harmony}, which combines hidden state representations from the model with token-level uncertainty to obtain a more comprehensive measure of reliability. In the future, these approaches could be further explored, as their inclusion may help mitigate issues related to inaccurate or ambiguous descriptions.




\section{Conclusions}
\label{sec:conclusions}
In this research, we introduce \method, a new model that presents a geographical invariant representation to mitigate performance loss caused by geographical domain shifts in camera trap image recognition.
\method~addresses geographical domain shifts by leveraging robust features extracted from the image and image-text encoder while also incorporating enhanced category descriptions generated by an LLM. 
Our extensive experiments demonstrate that \method~outperforms state-of-the-art models in camera trap image classification, particularly when there are geographical domain shifts between the training and testing datasets, all while preserving its open-vocabulary capabilities.

\bmhead{Data Availability}
Snapshot Serengeti data and Terra Incognita are publicly available on LILA BC at [https://lila.science/datasets/snapshot-serengeti] and on Caltech Camera Traps at [https://beerys.github.io/CaltechCameraTraps/].  For reproducibility, we use the preprocessed image data of Snapshot Serengeti as provided by WildCLIP [https://doi.org/10.1007/s11263-024-02026-6].

\bmhead{Acknowledgements}
This work was supported by Universidad de Antioquia - CODI (project 2024-73410), by the ANR (French National Research Agency) under the JCJC project DeSNAP (ANR-24-CE23-1895-01), and by the Academic Grant from NVIDIA AI.  

\begin{appendices}

\section{Prompts}
\label{supp: prompts}

In this section, we provide the prompts for the LLM and LLaVA used in \method.

\subsection{Prompt LLM}
The prompt used to generate the LLM description of the animal species follows a structured format based on the methodology described in \citep{pratt2023does} and is presented below.\\\\
\noindent\fbox{%
    \parbox{\columnwidth}{%
        You are an AI assistant specialized in biology and providing accurate and detailed descriptions of animal species. We are creating detailed and specific prompts to describe various species. The goal is to generate multiple sentences that capture different aspects of each species' appearance and behavior. Please follow the structure and style shown in the examples below. Each species should have a set of descriptions that highlight key characteristics.\\\\
        Example Structure:\\\\
        Badger: 
        \begin{itemize}
            \small
            \itemsep0em 
            \item a badger is a mammal with a stout body and short sturdy legs.
            \item a badger's fur is coarse and typically grayish-black.
            \item badgers often feature a white stripe running from the nose to the back of the head dividing into two stripes along the sides of the body to the base of the tail.
            \item badgers have broad flat heads with small eyes and ears.
            \item badger noses are elongated and tapered ending in a black muzzle.
            \item badgers possess strong well-developed claws adapted for digging burrows.
            \item overall badgers have a rugged and muscular appearance suited for their burrowing lifestyle.
        \end{itemize}
    }%
}
\subsection{Prompt LLaVA}
The prompt used in LLaVA follows the approach described in \citep{fabian2023knowledge} and is structured as follows:\\\\
\noindent\fbox{%
    \parbox{\columnwidth}{%
        \small
        \textbf{SYSTEM}: You are an AI assistant specialized in biology and providing accurate and detailed descriptions of animal species.\textbackslash n $\ll 
        \text{image} \gg$ \textbackslash n\\
        \textbf{USER}: You are given the description of an animal species. Provide a very detailed description of the appearance of the species and describe each body part of the animal in detail. Only include details that can be directly visible in a photograph of the animal. Only include information related to the appearance of the animal and nothing else. Make sure to only include information that is present in the species description and is certainly true for the given species. Do not include any information related to the sound or smell of the animal. Do not include any numerical information related to measurements in the text in units: m cm in inches ft feet km/h kg lb lbs. Remove any special characters such as unicode tags from the text. Return the answer as a single paragraph.
    }%
}

\section{Hyperparameter Search Space}
\label{supp: hyperparameter search}

To select the hyperparameters, we performed a Monte Carlo partitioning of the dataset. We generated three different partitions, each created using a different random seed. For each partition, a subset of the development set (training + validation data) was randomly assigned to the training and validation sets.

We then conducted a random search over a predefined hyperparameter space, testing different hyperparameter combinations. For each combination, we trained the model on all three partitions separately and computed the accuracy for each setting. To determine the final performance of a configuration, we calculated the mean accuracy and standard deviation across the three partitions. This process was repeated 30 times, and the best hyperparameter combination was selected based on the highest mean accuracy. The search space included the following hyperparameters:

\begin{itemize}
    \small
    \itemsep0em 
    \item \textbf{Batch size}: $ b \in \{128, 256\} $
    \item \textbf{Hidden dimension}: $ h \in \{253 + 60k \mid k \in \mathbb{Z}, 0 \leq k \leq 11\} $
    \item \textbf{Learning rate}: $ \eta \in \{0.01, 0.02, \dots, 0.09\} $
    \item \textbf{Momentum}: $ m \in \{0.80, 0.82, \dots, 0.98\} $
    \item \textbf{Number of epochs}: $ e \sim \mathcal{U}(25, 100) $ (randomly sampled between 25 and 100)
    \item \textbf{Temperature ($\tau$)}: $ \tau \in \{0.1, 0.01, 0.001\} $
    \item \textbf{$\alpha$}: $ \alpha \in \{0.4, 0.5, 0.6\} $
\end{itemize}

To evaluate the robustness of the selected hyperparameter combination, we further examined its consistency by computing the standard deviation in test accuracy across $100$ different random seeds for the out-of-domain experiments and $3$ different random seeds for in-domain experiments, due to the fact that in this experiment, we unfreeze the image encoder and fine-tune it.

\section{Templates}
\label{supp: templates}
In this section, we provide examples of templates specifically designed for the camera trap image recognition task. These templates are adapted from the ImageNet templates used in CLIP \citep{radford2021learning} and are presented below:
\begin{itemize}
    \small
    \itemsep0em 
    \item a photo captured by a camera trap of a \{\}.
    \item a camera trap photo of the \{\} captured in poor conditions.
    \item a cropped camera trap image of the \{\}.
    \item a camera trap image featuring a bright view of the \{\}.
    \item a camera trap image of the \{\} captured in clean conditions.
    \item a camera trap image of the \{\} captured in dirty conditions.
    \item a camera trap image with low light conditions featuring the \{\}.
    \item a black and white camera trap image of the \{\}.
    \item a cropped camera trap image of a \{\}.
    \item a blurry camera trap image of the \{\}.
    \item a camera trap image of the \{\}.
    \item a camera trap image of a single \{\}.
    \item a camera trap image of a \{\}.
    \item a camera trap image of a large \{\}.
    \item a blurry camera trap image of a \{\}.
    \item a pixelated camera trap image of a \{\}.
    \item a camera trap image of the weird \{\}.
    \item a camera trap image of the large \{\}.
    \item a dark camera trap image of a \{\}.
    \item a camera trap image of a small \{\}.
\end{itemize}
For each template, we replace ``\{ \}" by the specific category in $\mathcal{C}^D$.




\end{appendices}

\bibliographystyle{apalike}
\bibliography{egbib}

\end{document}